\documentclass[preprint,5p,twocolumn]{elsarticle}

\usepackage{amssymb}
\usepackage{amsmath}
\usepackage{lineno}
\usepackage{float} 

\usepackage{algorithmic}
\usepackage[utf8]{inputenc}
\usepackage{threeparttable}
\usepackage{graphicx}
\usepackage{algorithm,algorithmic}
\usepackage{hyperref}
\hypersetup{hidelinks}
\usepackage{textcomp}
\usepackage{textcomp}
\usepackage{tabu}
\usepackage{array}
\usepackage{multirow}
\usepackage{tabularray}
\usepackage{booktabs}
\usepackage{threeparttable}
\usepackage[table,x11names]{xcolor}
\usepackage{caption}

\journal{Computer Methods and Programs in Biomedicine}

\newcommand{\edit}[1]{{#1}}
\newcommand{\Edit}[1]{{#1}}

\begin{document}

\begin{frontmatter}



\title{Facial Surgery Preview Based on the Orthognathic Treatment Prediction}
\author[1,7]{Huijun Han}
\author[2]{Congyi Zhang\corref{cor1}}
\ead{cyzhang@cs.hku.hk}
\author[3]{Lifeng Zhu}
\author[1]{Pradeep Singh}
\author[6]{Richard Tai‑Chiu Hsung}
\author[6]{Yiu Yan Leung}
\author[2]{Taku Komura}
\author[7]{Wenping Wang}
\author[1]{Min Gu\corref{cor1}}
\ead{drgumin@hku.hk}

\affiliation[1]{organization={Discipline of Orthodontics, Faculty of Dentistry, the University of Hong Kong}, 
            city={Hong Kong SAR},
            country={China}}
\affiliation[2]{organization={Department of Computer Science, Faculty of Engineering, the University of Hong Kong}, 
            city={Hong Kong SAR},
            country={China}}
\affiliation[3]{organization={
School of Instrument Science and Engineering, Southeast University}, 
            city={Nanjing},
            country={China}}
\affiliation[6]{organization={Discipline of Oral and Maxillofacial Surgery, the University of Hong Kong}, 
city={Hong Kong SAR},
country={China}}
\affiliation[7]{organization={Department of Computer Science and Engineering, Texas A\&M University}, 
city={Texas},
country={USA}}

\cortext[cor1]{corresponding author}

\vspace{-15mm}
\begin{abstract}
\textit{Background and Objective:} Orthognathic surgery consultations are essential for helping patients understand how their facial appearance may change after surgery. However, current visualization methods are often inefficient and inaccurate due to limited pre- and post-treatment data and the complexity of the treatment. This study aims to develop a fully automated pipeline for generating accurate and efficient 3D previews of postsurgical facial appearances without requiring additional medical images. 

\textit{Methods:} 
The proposed method incorporates novel aesthetic criteria, such as mouth-convexity and asymmetry, to improve prediction accuracy. To address data limitations, a robust data augmentation scheme is implemented. Performance is evaluated against state-of-the-art methods using Chamfer distance and Hausdorff distance metrics. Additionally, a user study involving medical professionals and engineers was conducted to evaluate the effectiveness of the predicted models. Participants performed blinded comparisons of machine learning-generated faces and real surgical outcomes, with McNemar's test used to analyze the robustness of their differentiation.

\textit{Results:} Quantitative evaluations showed high prediction accuracy for our method, with a Hausdorff Distance of 9.00 millimeters and Chamfer Distance of 2.50 millimeters, outperforming the state of the art. Even without additional synthesized data, our method achieved competitive results (Hausdorff Distance: 9.43 millimeters, Chamfer Distance: 2.94 millimeters). Qualitative results demonstrated accurate facial predictions. 
The analysis revealed slightly higher sensitivity (54.20\% compared to 53.30\%) and precision (50.20\% compared to 49.40\%) for engineers compared to medical professionals, though both groups had low specificity, approximately 46\%. Statistical tests showed no significant difference in distinguishing Machine Learning-Generated faces from Real Surgical Outcomes, with p-values of 0.567 and 0.256, respectively. Ablation tests demonstrated the contribution of our loss functions and data augmentation in enhancing prediction accuracy.

\textit{Conclusions:} This study provides a practical and effective solution for orthognathic surgery consultations, benefiting both doctors and patients by improving the efficiency and accuracy of 3D postsurgical facial appearance previews. \edit{    The proposed method has the potential for practical application in pre-surgical visualization and aiding in decision-making.}
\end{abstract}

\begin{graphicalabstract}
\includegraphics[width=\textwidth]{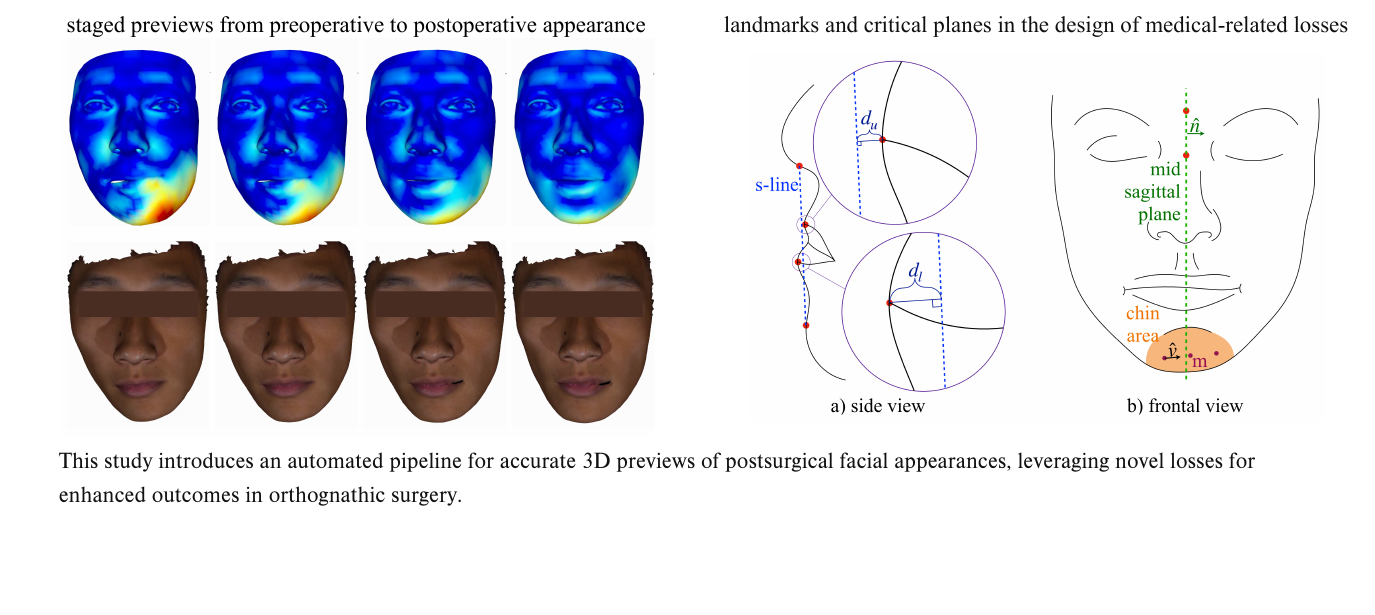}
\end{graphicalabstract}

\begin{highlights}

\item \edit{ We propose an automated method for 3D post-surgical prediction using only multi view images.}

\item \edit{We integrate mouth convexity and asymmetry criteria to enhance orthognathic planning.}

\item \edit{ We generate synthetic data from real cases to enhance training, improving robustness.}

\end{highlights}

\begin{keyword}
Computer-aided detection and diagnosis \sep geometric deep learning \sep visualization
\end{keyword}

\end{frontmatter}

\section{Introduction}
\label{sec:introduction}

Orthognathic surgery addresses facial asymmetry and abnormalities, significantly improving aesthetics, oral function, and psychosocial well-being. Despite these benefits, uncertainty about the appearance of the postoperative period often leads to anxiety among patients, affecting their decision-making process and communication with doctors. \edit{Visualizing expected surgical outcomes has emerged as a crucial step in mitigating presurgical anxiety, setting realistic expectations, and improving overall patient satisfaction, particularly when CBCT data is not available.}

Data-driven approaches have shown promise in assessing surgical necessity and complexity. For example, neural networks have been used to predict the need for orthognathic surgery from facial photographs~\cite{jeong2020deep} and to estimate the difficulty of tooth extraction from radiographic images~\cite{yoo2021deep}. However, these studies focus primarily on probability predictions rather than visual outcomes.

To address the need for predictive visualizations, researchers have begun exploring the use of machine learning algorithms to predict facial appearance after surgery~\cite{rekow2020digital}. 
These algorithms employ various advanced methods, including dense multilayer perceptrons (MLP)\cite{knoops2019machine,tanikawa:2021Aug:surplustreat,ter:2021Sep:three,chaiprasittikul:2023Jan:application,saeed:2023Aug:automatic,park2024does,KIM2023107853}, conditional generative adversarial networks (cGAN)\cite{Park:2022:orthodonticpred,laurinavivcius:2023March:improvement}, transformers~\cite{cheng2023prediction}, and convolutional neural networks (CNN)~\cite{ma2022machine}. These deep learning models excel at capturing spatial hierarchies in images, enabling refined feature extraction and accurate prediction of 3D coordinate changes.
Despite these advancements, existing methods remain clinician-oriented, heavily reliant on X-ray or CT data and anatomical landmarks provided by medical professionals~\cite{SANKAR2024}. This reliance limits full automation and increases costs, making these approaches inaccessible for patients seeking quick and intuitive consultations.
Moreover, several limitations persist. For example, the model in~\cite{ter:2021Sep:three} overlooks key reductions in facial asymmetry after surgery, potentially compromising accuracy for asymmetric patients. Similarly, Tanikawa’s network~\cite{tanikawa:2021Aug:surplustreat}, trained on fewer than 100 patients, faces overfitting issues due to high-dimensional input and output layers (18,000), while requiring CBCT data, which are challenging to obtain. Other approaches, such as Park’s cGAN~\cite{Park:2022:orthodonticpred}, are effective for orthodontic surgery but lack robustness for orthognathic procedures due to their complexity. Finally, methods like FC-Net~\cite{ma2021deep} require clinicians to provide precise bone movement data during inference, limiting their usability for direct patient engagement.

To accurately predict 3D post-surgery facial appearance, it is essential to use parametric 3D facial reconstruction methods that capture fine articulation, especially the jaw joint, which is critical for both aesthetic and functional outcomes in orthognathic surgery. Existing methods like Large Scale Facial Model (LSFM)~\cite{booth20163d} and Structure-Aware Editable
Morphable Model (SEMM)~\cite{ling2022structure} fail to model dynamic jaw movement, treating the face as a morphable structure where all features, including the jaw, are considered as a whole. SCULPTOR~\cite{qiu2022sculptor}, an articulated model, can capture skull-face joint distribution but is difficult to integrate due to limited data and incomplete code. To overcome this, we integrate FLAME~\cite{FLAME:SiggraphAsia2017,zheng:2023:flamebasedMultiView,liang_2024skulltofaceanatomyguided3dfacial}, a model with nonlinear jaw articulation, enabling vivid facial reconstruction and efficient encoding for accurate post-surgical predictions.

Our algorithm is designed to predict 3D post-surgery facial appearance based on preoperative facial scans of patients without the need for additional medical images (e.g., X-Ray or CBCT). To improve the robustness of the model, we develop a data augmentation method that substantially amplifies the available dataset for orthognathic treatment. To explicitly encode the assessment rules in orthognathic treatment, we introduce two medical losses, mouth-convexity loss and asymmetry loss, consistent with surgical definitions that help achieve the optimization goals of surgery.

In summary, our \textbf{contributions} are:
    (1) We integrate two criteria—minimizing mouth-convexity and minimizing asymmetry—into the machine learning procedure, which align with the goals of orthognathic surgery, to enhance prediction accuracy;
    (2) we devise an augmentation method to expand the available dataset, achieving more robust predictions;
    (3) We complete a fully automated pipeline that showcases the postoperative changes in facial appearance to patients in 3D.
 
\section{Methods}

\begin{figure*}[h]
 \centering 
 \includegraphics[width=\linewidth]{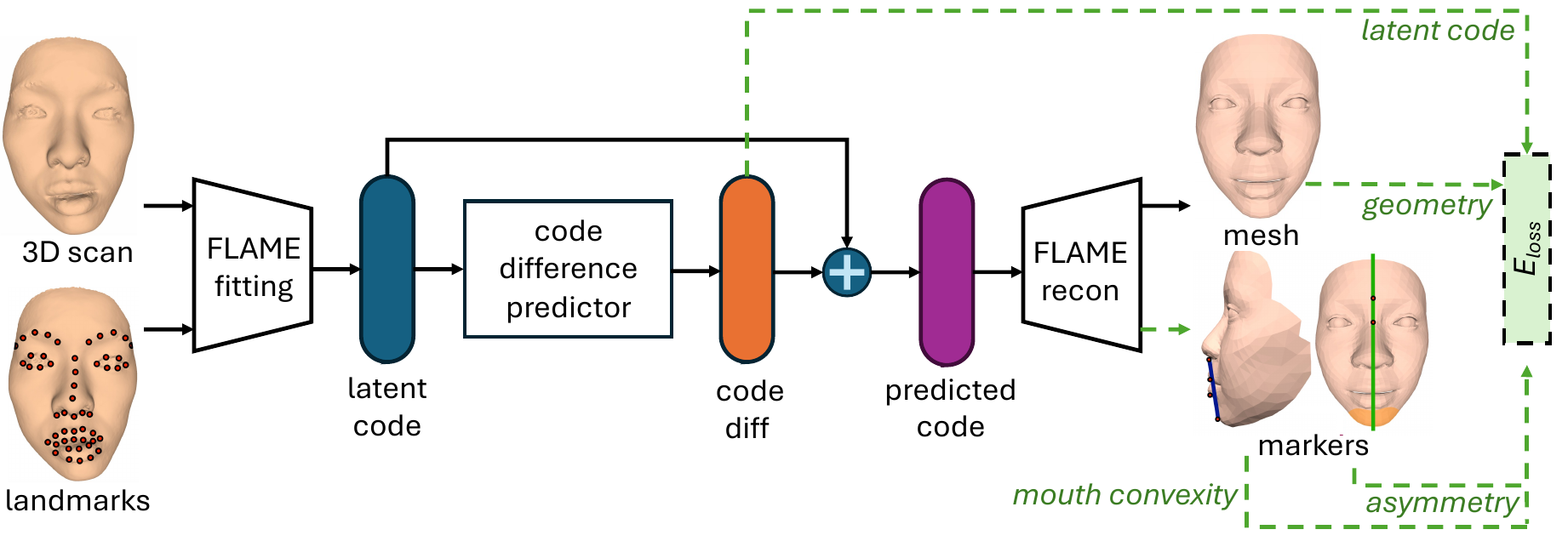}
    \vspace{-15pt}
 \caption{Net architecture for predicting the postoperative appearance from a captured 3D scan. During the training phase, the captured mesh and auto-annotated landmarks are first passed through the FLAME fitting procedure, where they are transformed into a compressed latent code. 
 It then passes through a predictor and has the code difference added to it. The FLAME reconstructing procedure helps to calculate well-designed losses using markers and mesh. With the help of medical, latent code, and geological types of loss, the parameter of the code difference predictor can be continuously updated. During the testing phase, the data do not flow through the dashed line. The predicted postoperative appearance is generated through reconstruction using FLAME.}
 \label{fig:pipelineForPrediction}
\end{figure*}

We present a simple and effective pipeline to predict 3D face models after surgery, as shown in Fig.\ref{fig:pipelineForPrediction}. 
\edit{
The reconstructed 3D facial model is initially encoded into a latent code using FLAME~\cite{FLAME:SiggraphAsia2017}, a parametric facial model. A face predictor then estimates the difference between post- and pre-surgery facial features, guided by custom-tailored loss functions that specifically incorporate facial asymmetry and mouth-convexity rules relevant to orthognathic surgery (explained in Section 2.1). These rules are directly applied to the predicted facial geometry and back-propagated through the network in a differentiable manner (discussed in Section 2.3). To provide a fully textured post-treatment preview, we leverage the mesh coherence of the FLAME model, transferring the texture from pre-treatment 3D facial scans to the predicted face models (explained in Section 2.4). 
To address the challenge of limited training data, we introduce a novel data augmentation technique that significantly increases the number of face pairs, enhancing the robustness of the model (described in Section 2.2). 
The effectiveness of the predicted models is evaluated through a user study involving both medical professionals and engineers, with the robustness of the predictions assessed using McNemar’s test (outlined in Section 2.5).
}

\subsection{Prediction network}

\begin{figure}[!h]
 \centering 
 \includegraphics[width=\linewidth]{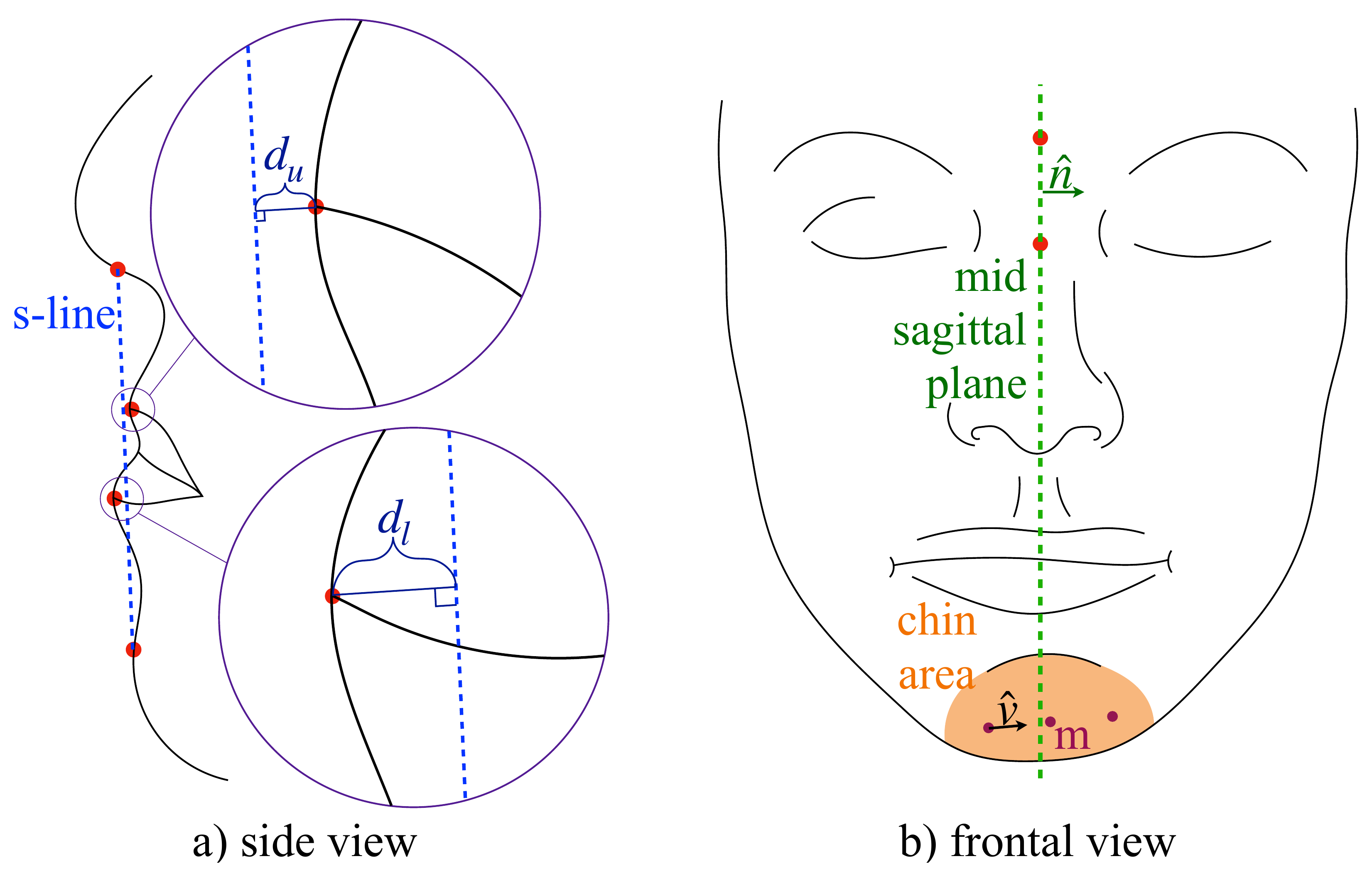}
    \vspace{-10pt}
 \caption{ Side view (left) and frontal view (right) of a orthognathic patient for calculating mouth-convexity loss and asymmetry loss respectively.}
 \label{fig:protrude}
\end{figure}

We adopt a predictor and its associated losses for the prediction of facial appearance of orthognathic surgery, as shown in Fig. \ref{fig:pipelineForPrediction}. As a parameterized and differentiable facial model, FLAME can serve as both a pre-trained encoder, compressing geometric information into a latent code, and also a decoder layer, providing the necessary geometric information in a differentiable manner for explicit supervision, which helps update the parameters in the code difference predictor.

The overall loss function used in our algorithm is composed of four components: the mouth convexity loss, the asymmetry loss, the latent code loss, and the geometry loss. The main challenge in training the predictor comes from the imperfect training data. Our training data consist of real orthognathic surgery cases. However, due to some practical limitations or customized considerations, the post-treatment facial appearance may not be ideal from a clinical perspective, e.g. some of them are still asymmetric or protruding to a certain extent. Adopting a pure data-driven loss in supervision would yield the same artifacts as observed in the training data. To this end, we formulate two explicit clinical rules as our novel losses, the mouth-convexity loss and the asymmetry loss, to enhance the functional and aesthetic aspects.

\textbf{Mouth-convexity loss:} 
\emph{Mouth convexity} refers to a measurement used in orthognathic surgery to describe the relative position of the mouth, nose, and chin. Mouth convexity helps categorize facial profiles into two main types: convex profiles (where the lower jaw protrudes outward, creating a rounded appearance) and concave profiles (where the lower jaw appears retruded or inwardly curved, resulting in a flatter facial contour). 

Orthognathic surgery is a highly effective approach to correcting mouth convexity. By carefully realigning the jawbones, this surgical procedure can significantly improve both the functional and aesthetic aspects of the patient's facial structure. Therefore, we tailor a specific loss function that penalizes the protruding mouth issues, so as to enable the network to generate more pleased facial geometries.

According to \cite{STEINER1960721}, a reference line known as \emph{Steiner-line} (s-line, blue dashed line in Fig.\ref{fig:protrude}.a) can be drawn from the middle of the nose base to \emph{pogonion} (the extreme anterior point of the chin) to serve as a basis for assessing the protrusion of the mouth. 
We denote the distances from the s-line to the midpoints of the upper and lower lips as $d_u$ and $d_l$ respectively.
Medical standards suggest that lip midpoints within a range of 3 millimeters deviation from the s-line are acceptable. To this end, we devise a mouth convexity loss function $L(d)$, where the squared distance is used to penalize cases with mouth convexity.
\begin{equation}
L(d) =
\begin{cases}
0, & \text{if } d < 3mm \\
(d-3)^2, & \text{if } d \geq 3mm
\end{cases}
\end{equation}
Mouth-convexity loss $L_{p}$ is the sum of $L(d_u)$ and $L(d_l)$.

\textbf{Asymmetry loss:} The degree of chin asymmetry in orthognathic surgery is a major concern, and can be quantified by measuring the symmetry of the chin with respect to the \emph{mid sagittal plane} \cite{dobai2018landmark} shown in Fig.\ref{fig:protrude}.b, which is the focus of our designed asymmetry loss.

According to the medical definition, we determine the mid-sagittal plane $S$ (green dashed line in Fig.\ref{fig:protrude}.b) by solving a plane that passes through points at the midpoint of the eyebrows and the midpoint of the inner corners of the eyes (red points in Fig.\ref{fig:protrude}.b).
Because all 3D reconstructed head models of patients are captured with the same natural head position, the ideal unit normal vector $\hat{\mathbf{n}}$ of the mid-sagittal plane is always along the $x$-axis.
In practice, for each case, we solve the least squares system to determine the mid-sagittal plane $S$.

As FLAME uses a topologically symmetric template mesh, we first pair all vertices in the chin area, denoted as $\{p_i, q_i\}_{i=1,\dots,k}$ (orange region in Fig.\ref{fig:protrude}.b). Then we compute the unit direction vectors $\hat{v}_i$ of the line segments connecting the paired points $p_i$ and $q_i$ and their midpoints $m_i=\frac{1}{2}(p_i+q_i)$. Following that, we can measure the overall asymmetry of paired points with respect to the mid-sagittal plane using the asymmetry loss
\begin{equation}
    L_a = \sum_{i=1}^k {d(m_i,S)} + (1- \hat{n}\cdot\hat{v}_i)
\end{equation}
where $d(m,S)$ denotes the distance function from point $m$ to mid-sagittal plane $S$ and $\cdot$ is the dot product.

\textbf{Latent code loss:} With the aid of a parameterized model serving as an encoder, each pair of preoperative and postoperative captured scans can be encoded as a pair of latent codes. To improve the approximation between the predicted and ground-truth values in the latent space during the training phase, we calculate the squared error of the latent codes as our latent code loss $L_f$:
\begin{equation}
    L_f = \| \vec{\beta}_{pred} - \vec{\beta}_{gt} \|_2^2
\end{equation}
where $\| \cdot \|_2$ denotes the $l^2$ norm, $\Vec{\beta}_{pred}$ and $\vec{\beta}_{gt}$ are the latent codes of predicted and postoperative GT faces.

\textbf{Geometry loss:} The geometry loss $L_g$ consists of two parts: the point-to-point distance and surface normal errors between the predicted and true meshes: 
\begin{equation}
   L_g = \frac{1}{N}\sum_{i=1}^N \| p_i^{pred} - p_i^{gt} \|_2^2  + w\frac{1}{M} \sum_{j=1}^M 1-cos(\theta_j), 
\end{equation}
where $N$ and $M$ are the numbers of points and triangles in the face region, $p^{pred}$ and $p^{gt}$ are the points on the predicted and ground-truth mesh, $\theta$ is the angle between the predicted and ground-truth surface normal, and $w$ is the balancing weight.

The geometry part is designed to encourage all the points within our facial mask to be close to their ground-truth values, preventing solely focusing on the chin and mouth areas due to the mouth-convexity loss and asymmetry loss. In addition, the normal part helps the network distinguish between the upper and lower lips, aiding in a better understanding of the facial appearance.
We conducted thorough ablation studies in Section \ref{sec::res:abla} to verify that geometry loss and latent code loss are not redundant.

\textbf{Overall loss function:} we compute the weighted sum of the four losses to form the total loss:
\begin{equation}
    L =  \alpha_{p} L_{p} + \alpha_{a} L_{a} + \alpha_{f} L_{f} +\alpha_{g} L_{g}.
\end{equation}

\subsection{Data augmentation}

A sufficient amount of diverse data is crucial for enhancing the robustness of machine learning models \cite{zhang2021understanding}. In orthognathic surgery, data augmentation offers significant advantages by improving the algorithm's ability to generalize across a wide range of cases. By simulating diverse populations, data augmentation enhances the model's predictive accuracy and ensures reliable 3D facial outcome predictions.

A common assumption in orthognathic surgery is that there is a horizontal plane dividing the unchanged upper part of the face (above the plane) from the key modified region (below the plane), where the surgical alterations occur. The goal of our data augmentation method is to generate synthetic data where the upper face remains intact, preserving its natural characteristics above the horizontal plane, while the lower part, which is affected by the surgery, is modified according to the surgical requirements.
Traditional 3D data augmentation techniques, such as translation, rotation, and scaling, may not be effective in our framework, as they do not generate meaningful geometry that meets the requirements for preserving the natural structure and characteristics of the orthognathic surgery.

\begin{figure}
    \centering
    \includegraphics[width=0.95\linewidth]{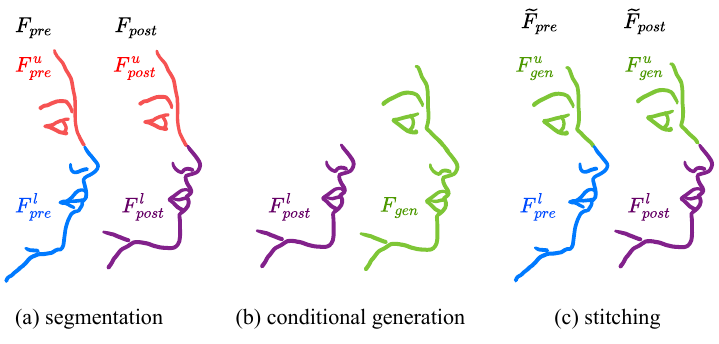}
    \vspace{-10pt}
    \caption{An illustration of synthetic data generation process. The process involves three stages: face segmentation, conditionally generating a synthetic face \( F_{gen} \), and stitching the upper and lower regions to create plausible pre- and post-surgery pairs for training.}
    \label{fig:aug}
\end{figure}

To create the synthetic data, we first identify the horizontal plane \( P_{s} \) that separates the unaltered upper face from the altered lower region. As illustrated in Figure~\ref{fig:aug}, for the pre-surgery scan, we obtain \( F_{pre}^{u} \) (upper part) and \( F_{pre}^{l} \) (lower part). For the post-surgery scan, we obtain \( F_{post}^{u} \) (upper part) and \( F_{post}^{l} \) (lower part). Note that \( F_{pre}^{u} \) is the same as \( F_{post}^{u} \).
Next, we conditionally generate a synthetic face \( F_{gen} \) based on the lower part of the postoperative scan \( F_{post}^{l} \) with a random variable .
Then, we use the horizontal plane \( P_{s} \) to separate the upper part of \( F_{gen} \), denoted as \( F_{gen}^{u} \).
Finally, we create the pair of generated faces \( \{ \Tilde{F}_{pre}, \Tilde{F}_{post} \}\) by stitching \( F_{gen}^{u} \) with both \( F_{pre}^{l} \) and \( F_{post}^{l} \), respectively.
\edit{We use a random variant 
$\xi$ to alter the latent code of the upper face, and then we use the decoder to reconstruct the facial geometry. }
We randomize the $\xi$ to synthesize multiple plausible pairs of \( \{\Tilde{F}_{pre} , \Tilde{F}_{post} \} \) to train the prediction network.
This approach ensures that the synthetic data reflects the surgical changes below the horizontal plane while keeping the upper part of the face unchanged.

\edit{In summary, our goal is to generate paired facial scans that reflect the pre- and post-surgery appearance. We modify the upper part of the face while directly copying the lower part from the dataset, following the paired changes between the pre- and post-surgery scans. As a result, we obtain a pair of faces where the upper part remains consistent, and the lower part exhibits the surgical changes. }

Additionally, we include a data cleaning step to remove outliers introduced by the fitting process. By calculating the fitting error at key facial landmarks and applying a threshold, we ensure that only meshes with an acceptable error are retained, thereby guaranteeing the accuracy and reliability of the synthetic meshes.
\Edit{To evaluate the impact of data augmentation, we conducted an ablation study, with the results presented in the last row of Table  \ref{tab:ablation} and the last row of Figure  \ref{fig:Augmentation}}

\subsection{Integrating FLAME as an encoder-decoder}

In our pipeline, FLAME is integrated not only as an encoder to obtain the latent code of captured scans, but also as a decoder to reconstruct facial meshes. As a full-head model with an underlying face skeleton tree composed of neck, jaw, and eyeball joints, FLAME model\cite{FLAME:SiggraphAsia2017} is defined as:
\begin{equation}
    M(\vec{\beta},\vec{\theta},\vec{\psi}) = W(T_P(\vec{\beta},\vec{\theta},\vec{\psi}),\mathbf{J}(\vec{\beta}),\vec{\theta},\mathcal{W}),
\end{equation}
where $\vec{\beta}$, $\vec{\theta}$ and $\vec{\psi}$ denote shape, pose and expression vector respectively. During our fitting and reconstruction procedure, the expression vector $\vec{\psi}$ is not involved, because the patient scanning data were obtained when they were relaxed and in neutral facial expressions and poses.

\subsection{Visualization of prediction as a textured mesh}
\label{sec:deformation}

We aim to provide individuals seeking surgical consultations with a textured 3D mesh that can be viewed from multiple angles, offering a comprehensive and intuitive visualization of their postoperative appearance. 

To achieve this, we deform the textured scan to align with the geometry of the predicted parametric model.  
Specifically, using barycentric coordinates, we deform the vertices to match the geometry of the predicted model. 
For each vertex in the scanned mesh, the closest point in the parametric model is identified as its correspondence, and the displacement for each vertex is computed based on the barycentric coordinates \((\lambda_1, \lambda_2, \lambda_3)\) and their predicted  displacements \(d_1\), \(d_2\), and \(d_3\). The texture is then morphed using the formula \(\lambda_1 d_1 + \lambda_2 d_2 + \lambda_3 d_3\) to approach the geometry of the predicted parametric model.

\subsection{User study}
The aim of this study was to assess the participants' ability to differentiate between machine learning-generated faces (MLG) and real surgical outcomes (RSO). 
\edit{
Participants in this study consisted of five medical professionals (comprising two surgeons and three orthodontists) with a minimum of 5 years of experience in their respective fields. Additionally, 15 engineers (including twelve males and three females) were included in the study, each possessing a minimum of 3 years of relevant experience in engineering and technical disciplines. The selection criteria for medical professionals were based on their expertise in surgical procedures and orthodontic treatments, ensuring a minimum threshold of 5 years of clinical experience. Engineers were chosen for their technical background and familiarity with machine learning concepts, with a minimum of 3 years of professional experience in their respective fields. 
}
Preoperative and post-surgery 3D facial data were collected for each patient, standardized using FLAME parameters, and textures were applied consistently across all data. To facilitate participants' observation of facial symmetry and feature positioning, both frontal and side views of the faces were provided, along with a 180-degree rotating animation from left to right. The preoperative images were clearly labelled, while the MLG and RSO were presented to the participants in a blinded manner. Each participant was simultaneously shown two images (labelled as A and B) and asked to distinguish between MLG and RSO without knowing the group to which each image belonged. 
\edit{A total of 30 randomly selected images (including A and B) were presented to each participant. }
Participants were required to digitally record their responses. The responses provided by the participants were then compared with the correct answers separately for engineers and medical professionals, and specificity, sensitivity, and precision values were calculated for each group. To further analyze the participants' ability to differentiate between MLG and RSO, a McNemar’s test was conducted for each group.
\section{Results}

\subsection{Datasets}

A total of 163 pairs of pre- and post-operative 3D facial scans were collected using the 3dMD facial scan acquisition system from patients undergoing orthognathic surgery at the University of Hong Kong School of Dentistry. 
Static three-dimensional (3D) images of each participant were taken using a 3dMDface System (3dMD Inc., Atlanta, GA, USA) by professional photographers. 
The accuracy of the system had been previously published and was reported to be lower than 0.2 mm root mean square (RMS)~\cite{shan2021anthropometric}. 
The system was calibrated according to the manufacturer's instructions before each image capture. 
During scanning, patients were relaxed and instructed to look straight into a mirror at their own eyes, ensuring the capture of their forehead, chin, and ears. 
Immediately prior to image capture, participants were seated 100 cm away from the system, looking forward with the Frankfort plane parallel to the floor, and any glasses and jewelry were removed. 
The camera system captured six 2D images—four black-and-white pictures to depict facial structures and spatial relationships, and two colored images to project texture information onto the mesh framework. 
The scan process took 1.5 milliseconds and the 3D facial surfaces were exported as Wavefront OBJ files for further processing. 
Postoperative scans were recorded at least three months after the surgery to ensure that facial swelling had subsided, providing an accurate reflection of the patient's post-recovery appearance. 
The male-to-female ratio was 59:104, with the majority of patients being of Asian descent.

\subsection{Implementation details}

\subsubsection{Exclusion of non-Facial elements}
Data cleaning is a crucial step in our prediction algorithm, as it ensures that the focus remains solely on the relevant facial regions by deliberately excluding non-facial elements such as hair and disposable surgical caps. This process is essential to eliminate extraneous data that could introduce noise, allowing the algorithm to concentrate on critical facial features like the chin and nose. By removing these non-facial elements, we not only enhance the fitting accuracy but also improve the overall performance of the model. The careful segmentation and exclusion of non-facial data ensure that the algorithm's analysis is based on high-quality, relevant information, leading to more precise and reliable predictions.

To isolate the facial regions, we render the 3D scans from three distinct viewpoints: frontal and both side views with a 45-degree rotation. We then apply BiSeNet~\cite{yu2018bisenet}, a bilateral segmentation network, to segment the facial regions from these images, retaining only the pertinent areas for subsequent analysis. Once the 2D segmentation maps are obtained, we reproject them back into the 3D space to ensure that the segmented regions correspond accurately to the facial areas in the original 3D scans. The non-facial elements, identified in the segmentation, are then removed from the 3D scans, leaving only the relevant facial regions for further analysis.

\subsubsection{Landmarks annotation}
We utilized Supervision by Registration \cite{dong2018sbr} in conjunction with a facial landmark detector to automatically annotate facial landmarks. A point light was positioned in front of the patient's face to render the 3D mesh into a frontal view image. A pre-trained landmark detector was then applied to extract 2D coordinates for 68 facial landmarks. These 2D coordinates were subsequently transformed back to 3D using the vertex-to-pixel mapping. 
For challenging preoperative cases, any discrepancies or drift in the landmark positions were manually corrected by a professional to ensure the dataset's accuracy and reliability.

\subsubsection{Training settings} 

Our neural network comprises two fully connected modules with a hidden layer of $100$ dimensions and input and output layers of $300$ dimensions. Additionally, the modules are connected by a batch normalization layer, a nonlinear layer activated by ReLU, and a dropout layer with a $50\%$ probability. The balance parameters $\alpha_{p}$, $\alpha_{a}$, $\alpha_{f}$, and $\alpha_{g}$ are set to $5000$, $5000$, $1$, and $1$ respectively.

During the training process, we set the batch size to $150$, and trained our model for a total of $500$ epochs. The original learning rate was set to $10^{-3}$, and we employed a learning rate decay strategy. Specifically, we decayed the learning rate by $50\%$ every $100$ epochs, which helped to prevent over-fitting and improve the generalization ability of our model. We conducted our training on a \texttt{NVIDIA RTX 3090} GPU. 

The training process took approximately 25 minutes to complete.
To ensure the robustness and reliability of our results, we employed a 5-fold cross-validation strategy.
\edit{The 163 valid pairs of facial scans were subjected to random shuffle and then split into 5 consecutive folds for cross-validation.}
Each fold was used once as the validation set, while the remaining four folds formed the training set.
Data splitting was performed prior to data augmentation, which was applied exclusively to the training dataset. We report the average score in Table \ref{tab:Performance}.

\subsection{Analysis of user study results}

\begin{table}[h]
\centering
\captionsetup{font=small}
\caption{Comparison of Sensitivity, Specificity, and Precision values for Engineers and Medical Professionals in distinguishing Machine Learning-Generated faces (MLG) and Real surgical outcomes (RSO)}
\label{tab:ComparisonSSP}
\begin{tabular}{lcc}
\hline
 & Medical Professionals & Engineers \\
\hline
Sensitivity & 53.30\% & 54.20\% \\
Specificity & 45.30\% & 46.20\% \\
Precision & 49.40\% & 50.20\% \\
\hline
\end{tabular}

\end{table}

The sensitivity, specificity, and precision values are presented in Table \ref{tab:ComparisonSSP}. The results indicated that engineers had a slightly higher sensitivity percentage (54.20\%) compared to medical professionals (53.30\%), suggesting that engineers were slightly better at identifying Machine Learning-Generated faces (MLG) and Real Surgical Outcomes (RSO) accurately. However, both groups exhibited low specificity percentages, with engineers at 46.20\% and medical professionals at 45.30\%, indicating challenges in distinguishing between MLG faces and RSO. In terms of precision, engineers had a slightly higher percentage (50.20\%) compared to medical professionals (49.40\%), suggesting a slightly higher accuracy for engineers in identifying MLG faces. Nevertheless, there was no statistically significant difference in the ability to differentiate between MLG faces and RSO within each group, with p-values of 0.567 for engineers and 0.256 for medical professionals (Table \ref{tab:statistical}). 
\edit{
Additionally, Table \ref{table:confusionMatric} presents the confusion matrix, illustrating the classification performance for both groups.}
\Edit{Noteworthily, although the subjective measure employed in this study holds value in assessing participants' ability to differentiate between MLG faces and RSO, it is important to acknowledge that subjective evaluation alone may not entirely validate clinical accuracy. 
}

\begin{table*}[h!]
\centering
\begin{threeparttable}
\captionsetup{font=small}

\caption{{Comparison of Engineers' and Medical Professionals' ability to distinguish between Machine Learning-Generated faces and Real surgical outcomes}}
\label{tab:statistical}

\begin{tabular}{lcc}

\hline
 & \edit{Medical Professionals ($N_{med} = 5$) $n=150$ }& \edit{Engineers ($N_{eng}= 15$) $n=450$} \\
\hline
MLG & 34 (23\%) & 104 (23\%) \\
MLG identified as RSO & 41 (27\%) & 121 (27\%) \\
RSO identified as MLG & 35 (23\%) & 103 (23\%) \\
RSO & 40 (27\%) & 122 (27\%) \\

\edit{95\% CI} & \edit{12.58 to 17.02} &\edit{14.01 to 16.12} \\ 
P-value  & 0.567 &0.256  \\
\hline
\end{tabular}
\begin{tablenotes}
\footnotesize
\item[] \edit{$N_{med}$, Number of medical professionals; $N_{eng}$, Number of engineers; n, number of responses; }
\item[] MLG, Machine learning generated faces; RSO, Real surgical outcomes; \edit{CI, Confidence Interval;}
\item[] McNemar chi-square test was performed unless otherwise mentioned.
\item[*] $p < 0.05$ (in \textbf{bold} \textit{italics}), considered statistically significant.
\end{tablenotes}
\end{threeparttable}
\end{table*}

\begin{table*}[h]
    \centering
    \captionsetup{font=small}
    \begin{threeparttable}
 \caption{\edit{Confusion matrix presenting True Positives, False Positives, True Negatives, and False Negatives.}}
 \label{table:confusionMatric}
\begin{tabular}{clll}
\hline
\multicolumn{1}{l}{}                                                            &                    & \multicolumn{2}{l}{True Responses} \\ \hline
\multicolumn{1}{l}{}                                                            &                    & RSO              & MLG             \\ \cline{2-4} 
\multirow{4}{*}{\begin{tabular}[c]{@{}c@{}}Predicted \\ Responses\end{tabular}} & Predicted RSO\_med & 53.30\%          & 54.70\%         \\
                                                                                & Predicted MLG\_med & 46.70\%          & 45.30\%         \\
                                                                                & Predicted RSO\_eng & 54.20\%          & 53.80\%         \\
                                                                                & Predicted MLG\_eng & 45.80\%          & 46.20\%         \\ \hline

\end{tabular}
\begin{tablenotes}
\footnotesize
\item[]  RSO: Real surgical outcomes;
\item[]  MLG: Machine learning-generated faces.
\end{tablenotes}
\end{threeparttable}

\end{table*}

\subsection{Quantitative metrics}

We introduce two key metrics to quantitatively evaluate the accuracy of the predictions: the Hausdorff distance and the Chamfer distance. These metrics are commonly used to assess the similarity between two point clouds or surfaces, providing insights into the geometric alignment of the predicted and ground truth meshes.

The \textbf{Hausdorff distance} measures the greatest of all the distances from a point in one set to the closest point in the other set~\cite{576361}. It is defined as:
\begin{equation}
d_H(A, B) = \max \left( \sup_{a \in A} \inf_{b \in B} \|a - b\|, \sup_{b \in B} \inf_{a \in A} \|b - a\| \right)
\end{equation}
where \( A \) and \( B \) are two point sets, and \( \|a - b\| \) denotes the Euclidean distance between points \( a \) and \( b \). The Hausdorff distance is sensitive to outliers and emphasizes the maximum deviation between the two sets.

The \textbf{Chamfer distance} provides a more balanced measure of the overall geometric difference between two point clouds~\cite{8578127}. It is computed as:
\begin{equation}
d_C(A, B) = \frac{1}{|A|} \sum_{a \in A} \min_{b \in B} \|a - b\|^2 + \frac{1}{|B|} \sum_{b \in B} \min_{a \in A} \|b - a\|^2
\end{equation}
where \( |A| \) and \( |B| \) are the number of points in sets \( A \) and \( B \), respectively, and \( \|a - b\| \) is the Euclidean distance between points \( a \) and \( b \). The Chamfer distance measures the average distance between the points in both sets, offering a less sensitive alternative to the Hausdorff distance for assessing surface matching.

Both of these metrics allow for a precise comparison of the predicted and actual 3D facial geometries, providing valuable insights into the performance of our model.

\begin{figure*}[!h]
    \centering
    \includegraphics[width=\textwidth]{figures/comparison-clean.pdf}
    \vspace{-10pt}
    \caption{\edit{Comparison of our method with LARS across four patient cases. On the left, the ground truth (GT) pre- and post-surgical scans are shown for reference. The middle columns display our predicted results, and the right columns show the LARS model's predictions. To compare the prediction errors with the real outcomes, heatmaps are provided showing the error distribution across facial regions, with the error bars located in the bottom-right corner. }}
    \label{rescomparison}
\end{figure*}
\subsection{Quantitative comparison}

These examples compare the real postoperative appearances of patients with the prediction results from our algorithm and LARS, the state-of-the-art model from~\cite{knoops2019machine}. 
We quantitatively compared the prediction errors of our network with those of LARS using the Chamfer distance (CD) and Hausdorff distance (HD) metrics. Table \ref{tab:Performance} summarizes the results across different settings. 

With the full dataset (\(1330\) samples), our network achieved a mean HD of \(9.00 \, \text{mm}\) and a mean CD of \(2.50 \, \text{mm}\), outperforming LARS, which had a mean HD of \(9.68 \, \text{mm}\) and a mean CD of \(2.77 \, \text{mm}\). This demonstrates that our network not only reduces overall errors but also minimizes large deviations, as indicated by lower maximum values for both metrics.

When trained without synthesized data (\(133\) samples), our network still performed better than LARS, with a mean HD of \(9.43 \, \text{mm}\) and a mean CD of \(2.94 \, \text{mm}\), compared to LARS's mean HD of \(9.67 \, \text{mm}\) and mean CD of \(2.98 \, \text{mm}\). This highlights the effectiveness of our approach even in limited data scenarios, though the inclusion of synthesized data further improves performance by enhancing the network's generalization capability. Additionally, we observed that data augmentation was very helpful in improving our performance, but had little effect on the performance of LARS.

\edit{
Table \ref{tab:statistical_results} presents the results of the statistical analysis using a t-test to compare our model with the LARS model based on two distance metrics: Hausdorff and Chamfer distances. The t-statistics and corresponding p-values indicate that both metrics show statistically significant differences between the models. Specifically, for the Hausdorff distance, our model demonstrated a t-statistic of 2.113 with a p-value of 0.039, which is below the 0.05 threshold for significance. Similarly, for the Chamfer distance, the t-statistic was 2.143 with a p-value of 0.036, also demonstrating a significant difference. These results suggest that our model outperforms the LARS model in terms of both distance metrics.}

\begin{table*}[!h]
\centering
\begin{threeparttable}
\captionsetup{font=small}
\caption{Performance Comparison}
\label{tab:Performance}
\begin{tabular}{lccccccc} 
\hline
\multirow{2}{*}{Algorithms}                                           & \multicolumn{3}{c}{HD* (mm) $\downarrow$} & \multicolumn{3}{c}{CD* (mm)~$\downarrow$} & \multirow{2}{*}{\begin{tabular}[c]{@{}c@{}}Data \\Amount\end{tabular}}  \\
& mean  & min   & max    & mean  & min   & max          &           \\ 
\hline
\textbf{OUR's  }                                                               & \textbf{9.00} & 7.63 &\textbf{ 11.30 }      & \textbf{2.50} & \textbf{1.24} & \textbf{3.60}        & \textbf{1330  }                                                                  \\
LARS                                                                  & 9.68 & 7.50 & 15.41       & 2.77 & 1.72 & 4.77        & 1330                                                                    \\
OUR's w/o synthesized data & 9.43 & 8.00 & 13.53       & 2.94 & 1.91 & 6.38        & 133                                                                     \\
LARS w/o synthesized data & 9.67 &\textbf{ 7.46} & 16.13       & 2.98 & 1.57 & 5.21        & 133                                                                     \\
\hline
\end{tabular}
\begin{tablenotes}

\footnotesize
\item [ *] HD and CD represent Hausdorff and Chamfer Distance respectively.
\end{tablenotes}
\end{threeparttable}
\end{table*}

\begin{table*}[ht]
\centering
\captionsetup{font=small}
\caption{\edit{Statistical analysis of Hausdorff and Chamfer distances.}}
\begin{tabular}{llccl}
\hline
\textbf{Comparison}        & \textbf{Metric}   & \textbf{t-statistic} & \textbf{p-value} & \textbf{Significance} \\ \hline
Our model vs LARS model & Hausdorff         & 2.113               & 0.039            & Significant          \\ \hline
Our model vs LARS model & Chamfer           & 2.143               & 0.036            & Significant          \\ \hline
\end{tabular}

\label{tab:statistical_results}
\end{table*}

\subsection{Qualitative Comparison}

To qualitatively evaluate our algorithm, we selected representative cases for visualization, as shown in Fig. \ref{rescomparison}. These examples compare the real postoperative appearances of patients with the prediction results from our algorithm and LARS ~\cite{knoops2019machine}. To ensure a fair assessment of accuracy, experiments were also conducted using the Basel network and compared against LARS.
The showcased patients underwent bimaxillary orthognathic surgery, with or without genioplasty.

For the first patient, our approach effectively predicted the correction of mouth alignment, showing only minor discrepancies in the lower face when compared to LARS. Notably, in the lower nasal region, the linear model predicted a longer, wider, and lower nasal base, deviating significantly from the surgical plan.
For the second patient, our algorithm successfully predicted the repositioning of the jaw, whereas LARS struggled to address the protruding chin accurately. 
In the cases of the third and fourth patients, both of whom underwent bimaxillary orthognathic surgery combined with genioplasty to shorten facial length by adjusting the chin’s tilt angle, our algorithm precisely captured the jaw’s angle changes, resulting in a more aesthetically balanced facial shape. In contrast, LARS failed to accurately predict these adjustments, underscoring the superior predictive capability of our method.

\section{Discussion}

\subsection{Ablation experiment}
\label{sec::res:abla}

\begin{figure}[tb]
    \centering
    \includegraphics[width=\columnwidth]{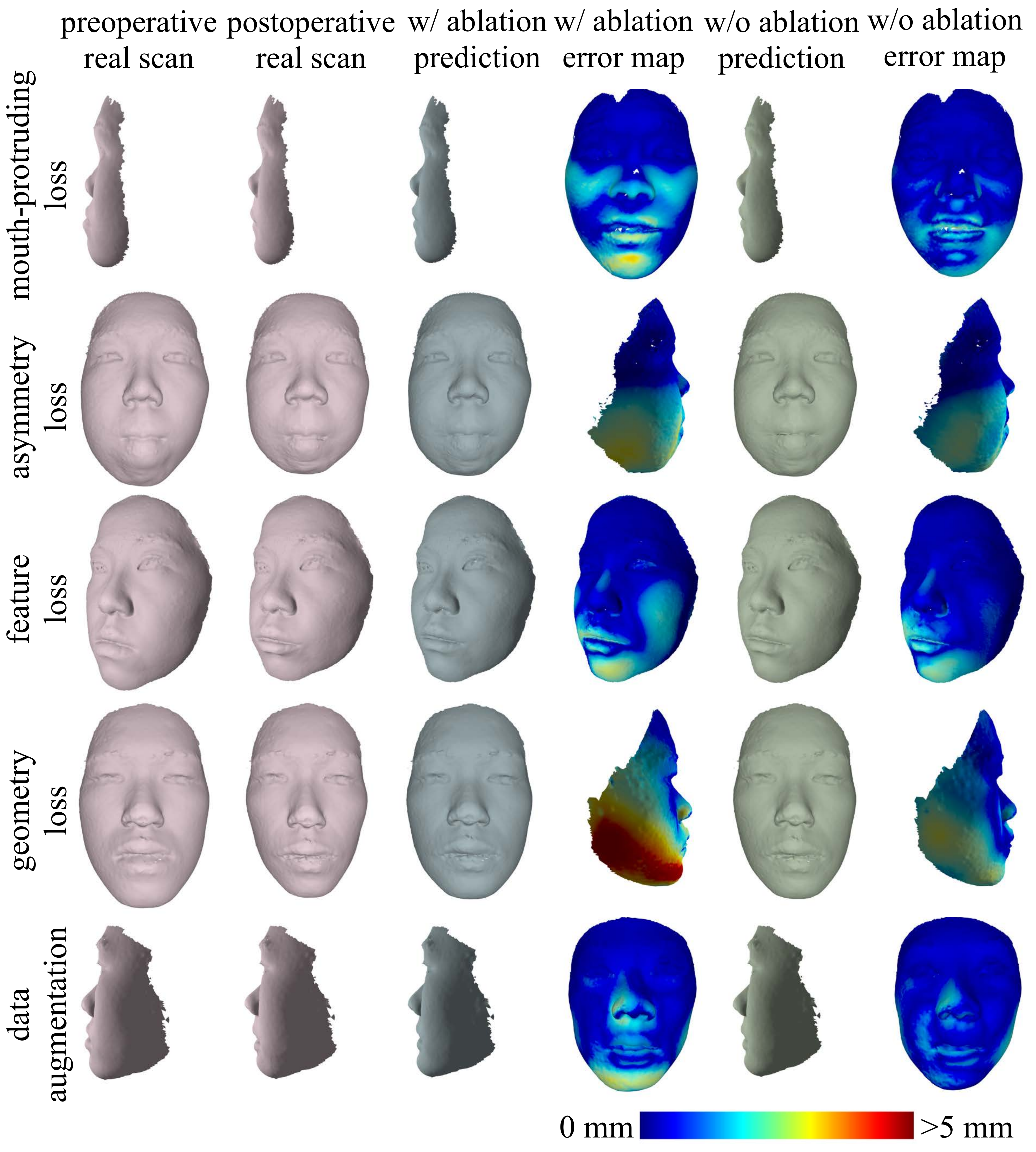}
    \vspace{-20pt}
    \caption{\edit{Impact of losses and data augmentation on predictions.
The left section illustrates actual surgical data, the middle section presents results after removing specific modules, and the right section showcases the full-model predictions.}}

    \label{fig:Augmentation}
\end{figure}

We conducted an ablation study to examine the effectiveness of the four losses and data augmentation techniques introduced in our model. \edit{Both quantitative (as shown in Table. \ref{tab:ablation}) and qualitative (as shown in Figure. \ref{fig:Augmentation}) results demonstrate their effectiveness.}

The ablation study highlights the contributions of various components in our model, as summarized in Table \ref{tab:ablation}. The full model (OUR's) achieves the best performance with a Hausdorff Distance of 8.999 mm and a Chamfer Distance of 2.503 mm. Removing the mouth-convexity loss slightly degrades performance (0.11\% in Hausdorff, 2.32\% in Chamfer), while the asymmetry loss has a more noticeable impact (1.59\% in Hausdorff, 2.72\% in Chamfer). The latent code loss affects face consistency (0.53\% in Hausdorff, 5.27\% in Chamfer), and the geometry loss is critical, causing the largest increases among the four losses (2.64\% in Hausdorff, 5.63\% in Chamfer).

Data augmentation plays a pivotal role in model training, as its removal causes the most pronounced effect, increasing Hausdorff Distance by 5.75\% and Chamfer Distance by 17.46\%, while reducing the training dataset by 90\%.\edit{ The addition of augmented data appears to help reduce prediction error.}

Figure \ref{fig:Augmentation} further underscores these findings, vividly illustrating the qualitative impact of each component. For instance, removing the mouth-convexity loss leads to an exaggerated chin, while the absence of the asymmetry loss results in pronounced facial asymmetry. Similarly, excluding the latent code loss causes notable drift in the cheek contours, and removing the geometry loss significantly exacerbates overall structural errors. 
\edit{
Additionally, data augmentation seems to play an important role, as its absence introduces noticeable inconsistencies in facial predictions.}

\begin{table}[!h]
  \centering
  \caption{Ablation study of our model}
  \label{tab:ablation}  
  \small
  \begin{tabular}{lccc}
    \hline

       Prediction Network & \multicolumn{1}{p{1.2cm}}{Hausdorff \newline Distance (mm)} &  \multicolumn{1}{p{1.2cm}}{Chamfer \newline Distance (mm)}  &  \multicolumn{1}{p{1.5cm}}{Training \newline Data Amount}\\
      
    \hline
    \textbf{OUR's} & \textbf{8.999} & \textbf{2.503 }&\textbf{ 1330} \\
    - Mouth-convexity loss & 9.009 & 2.561 & 1330 \\
    - Asymmetry loss & 9.142 & 2.571 & 1330 \\
    - Latent code loss & 9.047 & 2.635 & 1330 \\
    - Geometry loss & 9.237 & 2.644 & 1330 \\
    - Augmentated Data & 9.517 & 2.940 & 133 \\
    \hline
  \end{tabular}
\end{table}

\subsection{Evaluation of Methodology Robustness}
The results of our user study revealed that both engineers and medical professionals encountered similar challenges in accurately distinguishing between MLG faces and RSO. The minimal differences in sensitivity, specificity, and precision values between the two groups indicate that both groups faced similar difficulties, highlighting the reliability and accuracy of our novel method. The consistency in the performance of both groups confirms the effectiveness and robustness of our methodology and algorithm in predicting facial appearance following orthognathic treatment using only 3D face geometry.

\subsection{Comparison with commercial VSP tools}

Although Virtual Surgical Planning (VSP) tools (e.g., Dolphin, Morpheus 3D FaceMaker, and 3dMD Vultus) provide clinically accurate results for surgical planning, they require Computed Tomography (CT) or Cone Beam Computed Tomography (CBCT) imaging data as additional inputs~\cite{starch2023accuracy}. 

A patient-oriented solution for visualizing surgical outcomes should be more accessible than a professional doctor-oriented one. 
In the early consultation stage,  patients are not obligated to take CBCT or X-ray imaging, which would expose users to radiation. 
Instead, 3D facial reconstruction (e.g., 3dMD (3dMD Inc., Atlanta, GA, USA), Bellus3D (Bellus3D Inc., Los Gatos, CA, USA), and Ein-Scan3D (Shining 3D Technology, Hangzhou, China)) is much safer for post-treatment preview purpose.

\begin{figure}[!h]
  \centering
  \includegraphics[width=0.8\linewidth]{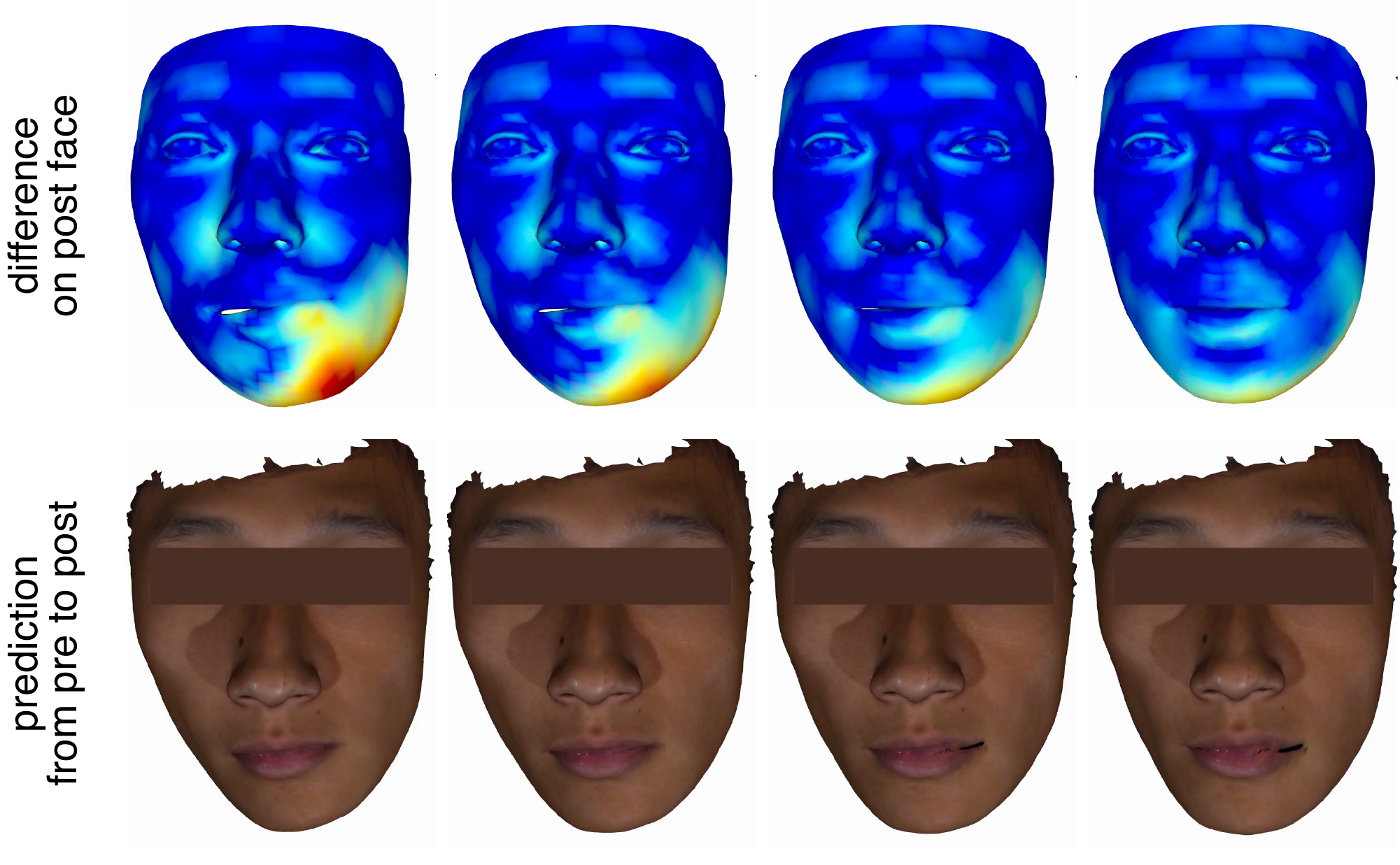}
  \vspace{-10pt}
  \caption{Visualization of the latent space interpolation results. The faces generated from interpolated latent codes are shown in the second row. The face on the far left is the pre-surgery face, and the face on the far right is the prediction of the post-surgery outcome. The first row represents the distances between these faces and the post-surgery face. Use the same color map as in Fig. \ref{fig:Augmentation}. }
  \label{fig:visualization}
\end{figure}

Our fully automated and highly efficient method delivers accurate visualizations, offering an automatic preview of facial surgery outcomes that serves as an effective educational and motivational tool for patients, particularly during the consultation phase. 
By functioning as a visual aid, it enhances patients' understanding and acceptance of treatment plans. To further engage patients, we generated an animation by interpolating the latent vectors, illustrating the transformation from pre-surgery to post-surgery. This visualization highlights the movement of the chin and the overall changes in facial appearance, as depicted in Fig. \ref{fig:visualization}. For a more comprehensive demonstration, please refer to our supplementary video.

\section{Limitation}
\Edit{

Despite the comprehensive analysis conducted in this study, several limitations need to be acknowledged. The current training dataset is limited to Asian patients, which may impact the generalizability of the findings. Therefore, future research endeavors can focus on expanding the dataset to encompass greater diversity, including various ethnicities for improved applicability across different populations. Additionally, the study may benefit from addressing long-term skeletal changes, as the dataset primarily consists of adults with post-surgical scans taken at least six months after surgery to capture stabilized facial structures. While results are deemed reliable over an extended period, ongoing monitoring and consideration of potential long-term changes will be crucial for enhancing the study's robustness. Furthermore, the model's inability to offer multimodal-based predictions for individual patient attributes, such as age, gender, and skin condition, underscores the need for additional data with multi-modal labels to enable more targeted predictions without compromising data integrity. The current focus on providing automatic visualization services for patients rather than adjustable parameters for orthognathic professionals highlights a potential limitation for medical professionals seeking customizable features for surgical planning. In future work, considerations may include incorporating user-friendly interfaces with adjustable parameters to cater to professional needs effectively. Moreover, while the model primarily supports facial prediction for Asian facial types, efforts to expand its applicability to other ethnic groups through further research and validation are essential. Finally, while the subjective evaluation provided insights into the perceptual differences between MLG faces and RSO, it is crucial to recognise the limitations of relying solely on subjective assessments. Subjective measures, by nature, may introduce bias and subjective interpretation, potentially affecting the overall validity and reliability of the study findings. To mitigate these limitations, future studies can integrate quantitative clinical metrics and objective measures to complement subjective assessments, enhancing the study outcomes' validity and reliability. Incorporating quantitative measures such as cephalometric analyses, facial landmark tracking, and patient-reported outcomes will facilitate a more comprehensive assessment and validation of machine learning-generated facial predictions in the context of orthognathic surgery.
}
\section{Conclusions}
In this paper, we introduce a novel method for predicting facial appearance following orthognathic treatment using only multi view images. During the training phase, our approach utilizes customized mouth-convexity and asymmetry losses, combined with latent code, geometric losses, and data augmentation, to enhance robustness and outperform existing methods in terms of accuracy.

\section*{Acknowledgments}
This work was approved by the IRB (UW21-140 HKU/HA HKW IRB) and funded by the GRF grants (17107321) from the RGC of Hong Kong. The authors declare no competing interests. The trial is registered under HKUCTR-2971.

\biboptions{sort&compress}
 \bibliographystyle{elsarticle-num-names}
 \bibliography{ref-short, ref}

\end{document}